\documentclass[conference, a4paper]{IEEEtran}
\IEEEoverridecommandlockouts

\ifCLASSINFOpdf
  \usepackage[pdftex]{graphicx}
  \DeclareGraphicsExtensions{.pdf,.jpeg,.png}
\else
\fi

\usepackage{cite}
\usepackage{comment}
\usepackage{amsmath,amssymb,amsfonts}
\usepackage{algorithmic}
\usepackage{subcaption}
\usepackage{graphicx}
\usepackage{textcomp}
\usepackage{booktabs}
\usepackage{multirow}
\usepackage[left=1.29cm, right=1.29cm, top=1.9cm, bottom=4.29cm]{geometry}

\usepackage{optidef}
\usepackage{xcolor}
\usepackage{hyperref}
\usepackage{longtable}
\usepackage{gensymb}
\usepackage[]{algorithm2e}

\usepackage{glossaries}
\usepackage{soul}
\usepackage{flushend}

 \makeglossaries
\usepackage[font=small]{caption}
\usepackage{enumitem}
\usepackage{anyfontsize}
\usepackage{fancyhdr}
\makeatletter
\renewcommand{\fnum@figure}{Figure \thefigure}
\makeatother

\title{On the Detection of Aircraft Single Engine Taxi using Deep Learning Models} 
\vspace{0.5cm}

\author{
\IEEEauthorblockN{Gabriel Jarry,  Philippe Very, Ramon Dalmau}
\IEEEauthorblockA{
\small
EUROCONTROL \\
Brétigny-Sur-orge, France\\
\{gabriel.jarry, philippe.very, ramon.dalmau\}@eurocontrol.int}

\and
\IEEEauthorblockN{Daniel Delahaye, Arthur Houdant}
\IEEEauthorblockA{
ENAC, Université de Toulouse\\
Toulouse, France\\
daniel.delahaye@enac.fr,  arthur.houdant@alumni.enac.fr} 
}

\begin{document}

\maketitle
\thispagestyle{fancy}

\begin{abstract}
The aviation industry is vital for global transportation but faces increasing pressure to reduce its environmental footprint, particularly CO2 emissions from ground operations such as taxiing. Single Engine Taxiing (SET) has emerged as a promising technique to enhance fuel efficiency and sustainability. However, evaluating SET's benefits is hindered by the limited availability of SET-specific data, typically accessible only to aircraft operators. In this paper, we present a novel deep learning approach to detect SET operations using ground trajectory data. Our method involves using proprietary Quick Access Recorder (QAR) data of A320 flights to label ground movements as SET or conventional taxiing during taxi-in operations, while using only trajectory features equivalent to those available in open-source surveillance systems such as Automatic Dependent Surveillance-Broadcast (ADS-B) or ground radar. This demonstrates that SET can be inferred from ground movement patterns, paving the way for future work with non-proprietary data sources. Our results highlight the potential of deep learning to improve SET detection and support more comprehensive environmental impact assessments.
\end{abstract}


\section{Introduction}
\label{sec
}

The aviation industry plays a vital role in global transportation, connecting people and economies across the world. However, it also accounts for approximately 2.5\% of global CO2 emissions, with this share expected to rise due to increasing demand for air travel~\cite{GOSSLING2020102194}. In addition to CO2 emissions, aviation contributes to non-CO2 effects, such as the formation of contrails, which can influence climate change~\cite{lee2021contribution}. 

Recognizing these environmental challenges, the aviation sector has committed to ambitious sustainability goals. Through initiatives like the U.S. NextGen program and the Single European Sky Air Traffic Management Research (SESAR) Joint Undertaking, the industry is working to reduce emissions significantly by 2035~\cite{undertaking2015european}. These efforts focus on a combination of advanced technologies, improved air traffic management (ATM) practices, and innovative operational procedures. For instance, techniques such as continuous descent operations (CDOs) and continuous climb operations (CCOs) have been successfully implemented to reduce fuel consumption and emissions during different in-flight phases~\cite{dalmau2015fuel, Clarke2004}.

In recent years, reducing emissions during ground operations has become a key focus of aviation's broader environmental strategy. Among the most promising techniques developed in this area is Single Engine Taxiing (SET), where aircraft taxi using only one engine instead of both. This approach has shown considerable potential for reducing fuel consumption and emissions during taxiing, as evidenced by studies at major airports such as London Heathrow, Orlando, and New York-LaGuardia~\cite{stettler2018impact, kumar2008analysis}. By decreasing fuel burn during ground operations, SET not only supports the industry's sustainability goals but also offers cost savings for airlines.

Despite its potential, fully evaluating the environmental benefits of SET is hampered by the lack of accessible data on its actual usage. SET is not automatically recorded and is usually implemented at the discretion of flight crews, making it difficult to systematically track. Although aircraft operators have access to proprietary Quick Access Recorder (QAR) data, which logs engine usage and can confirm SET, this data is typically not available for public use due to privacy and competitive concerns. This limitation creates a significant barrier to understanding the broader adoption and environmental impact of SET.

To address this issue, we propose a novel method for detecting SET operations that bypasses the need for proprietary data. Our approach leverages QAR data only to label historical ground operations as either SET (positive class) or conventional taxiing (negative class). Crucially, we use only ground trajectory information --features equivalent to those found in open data sources such as Automatic Dependent Surveillance-Broadcast (ADS-B) or radar -- as input to our model. By doing so, we demonstrate that SET operations can be inferred solely from aircraft ground movements, which opens the door for applying this method to widely available surveillance data in future implementations. 

In this study, we train a Convolutional Neural Network (CNN) on the labeled dataset to classify SET operations based on ground trajectories. While the QAR data is necessary to label the dataset and develop the model, the input features are chosen to reflect what can be captured by open surveillance systems like ADS-B or radar. This approach ensures that once the model is trained, it can operate without the need for proprietary data, making it scalable and accessible for broader use. This paper focuses on demonstrating the effectiveness of the model with ground trajectory data. Besides, we address the challenges of identifying the precise initiation of SETs.

By providing means to detect SET operations using open-source trajectory data, this research significantly contributes to the aviation industry's sustainability efforts. Our method not only facilitates comprehensive assessments of SET adoption but also helps airlines and airports enhance their operational efficiency and environmental practices, furthering their progress toward ambitious climate goals.

\section{State of the Art}
\label{sec:state}

This section presents the state of the art in two key areas. The first subsection reviews the latest advancements in fuel estimation techniques, while the second subsection explores the litterature around taxiing fuel flow estimation and single engine taxiing.

\subsection{Fuel Estimation}

In ATM, the accurate estimation of aircraft fuel consumption is still a major challenge. To address this challenge, researchers have developed a variety of modeling approaches, ranging from sophisticated neural networks and machine learning techniques to more conventional physical models. The most relevant work from each of these different approaches is highlighted in this subsection.

A well-known aircraft performance model that serves as the basis for Total Energy Modeling based aircraft trajectory simulation is EUROCONTROL's Base of Aircraft Data (BADA). BADA includes detailed fuel, drag, and thrust models for a wide range of aircraft types~\cite{nuic2010user, nuic2010bada}. OpenAP, an open source model created specifically for the research community to provide greater accessibility, encourages transparency and broad collaboration~\cite{sun2020openap}. To predict cruise fuel burn and performance characteristics for turbofan aircraft,~\cite{poll2021estimation, poll2021estimation2} recently proposed a method based on empirical data and aerodynamic theory.

Recent research has shown that machine learning is more effective than traditional physical models. For example, to improve model accuracy, \cite{chati2017gaussian} used operational data and Gaussian Process Regression (GPR). Using Classification And Regression Trees (CART) and least squares boosting, the same authors improved the accuracy of the emissions inventory by predicting fuel flow rates~\cite{chati2016statistical}. Similarly, \cite{baumann2020modeling} achieved better results than traditional fuel flow models by combining machine learning techniques with full-flight sensor data. Using sophisticated neural networks such as radial basis function networks,~\cite{baklacioglu2021predicting} demonstrated even greater accuracy in fuel flow prediction. By comparing neural network models with specific aircraft types,~\cite{trani_neural_2004} demonstrated how useful they could be for real-time simulation applications. Long-Short Term Memory (LSTM) neural networks were also used by \cite{li2021study} to accurately predict performance-based contingency fuel.

Without much experimental testing, \cite{kayaalp2021developing} achieved high accuracy with an LSTM model related to exhaust emissions and combustion efficiency. Similarly, \cite{metlek2023new} created a model that outperformed previous techniques in terms of accuracy in predicting aircraft fuel consumption. A unique method for predicting fuel consumption during different flight phases was introduced by \cite{baklacioglu2016modeling} through the integration of genetic algorithm-optimized neural networks. In addition, \cite{uzun2021physics} showed great promise in strengthening the resilience of neural network models by introducing a hybrid strategy that incorporates physics-based losses during training.

In our previous work~\cite{jarry2020approach}, we benchmarked neural networks for estimating on-board aircraft parameters such as fuel flow and flap during approach and landing, which can lead to better ATM metrics ~\cite{jarry2021toward}. In addition, we have provided an open-source generic aircraft fuel flow regressor for ADS-B data\cite{jarry2024towards}, which was trained using quick-access recorder data. Although this model can estimate fuel flow on the ground, it is unable to distinguish between single-engine taxiing and nominal operations.

\subsection{Taxiing and Single Engine}

Several studies have explored methods to reduce aircraft emissions and fuel consumption during ground operations, with Single Engine Taxiing (SET) emerging as a promising technique. Research at London Heathrow Airport highlights the significant reduction in fuel consumption and emissions when SET is used during taxiing. These benefits could be enhanced further by shortening the time before initiating SET \cite{stettler2018impact}. Similarly, another study shows how the implementation of single-engine taxi-out procedures at Orlando (MCO) and New York-LaGuardia (LGA) airports leads to notable reductions in pollutant emissions, by 27\% at MCO and 45\% at LGA \cite{kumar2008analysis}. These findings support the use of SET as a viable strategy to mitigate airport-related environmental impacts.

When considering overall taxi operations, fuel consumption and emissions vary significantly depending on the taxiing scenario. At Dallas/Fort Worth International Airport (DFW), stop-and-go taxiing causes a substantial 18\% increase in fuel burn, which indicates the need for more optimized taxiing procedures \cite{nikoleris2011detailed}. Similarly, a data-driven model based on Flight Data Recorder (FDR) data has been developed to estimate fuel burn during the taxi-out phase more accurately, accounting for factors such as taxi time and acceleration events. This model improves upon existing estimates by considering operational specifics, offering more precise predictions than the International Civil Aviation Organization (ICAO) standards \cite{khadilkar2012estimation}.

Further discrepancies in emissions estimates have been highlighted in a study comparing ICAO’s generalized parameters with actual data from Shanghai Hongqiao International Airport. This research shows that ICAO's predictions overestimate taxi-out times and emissions, underscoring the importance of using airport-specific data to enhance the accuracy of local air quality models \cite{xu2020characterizing}. Building on this, a refined model assessing fuel consumption and emissions during taxiing has been developed, which includes considerations such as low visibility and traffic conflicts. This model identifies electric green taxiing systems (EGTS) as the most effective solution for reducing emissions and fuel consumption \cite{zhang2019assessment}.

\subsection{Contribution of this paper}

In this paper, we propose data-based classification and regerssion eural networks that predict the use and the start of single engine taxis during taxiing operations. To the best of the author's knowledge, this is the first publication on the identification of single engine taxiing using ADS-B/radar-like parameters.

\section{Methodology}
\label{sec:process}

\subsection{Data collection and preparation}
\label{subsec:data}

\subsubsection{Flight Data and preprocessing}

The dataset consists of 16,452 A320 Quick Access Recorder (QAR) full flights with a sampling rate of one point per second from one airline. Flights land at several airports, the main one being Paris-Orly (31\%). The trajectories are truncated to the taxi-in phase to the gate (i.e. after landing when the aircraft leaves the runway). This feature is already computed in the QAR data, however, when using open-data, one can use the \texttt{$on\_taxiway$} function from the \emph{traffic}  library \cite{olive2019traffic}.

Since the target should be able to use such a model with either ADS-B or ground radar data, the model is trained only on features that would be available with such data. In addition, for each trajectory we compute both the 5s and 10s derivatives of ground speed and track angle. The different features are summarized in Table \ref{tb:features}. Finally, the trajectories are zero padded to match a consistent shape of (2048, 7).

To get a label (1 : SET, 0 no SET) we used the fuel flow information of the different engines. If we observe a fuel flow below 5 kg/h on one engine and above on the other, then that point is considered SET. In addition, we make sure that we observe at least 1 minute in this SET configuration to give the label 1.

The dataset is partitioned into training, validation, and test sets using an (80, 10, 10) repartition that ensures a similar positive and negative class distribution in all sets. The training set consists of 12445 negative samples and 714 positive samples. 

\begin{table}[h!]
\centering
\caption{Data features and units used to train the model}
\label{tb:features}
\begin{tabular}{ll}
\toprule
\textbf{Feature} & \textbf{Unit} \\
\midrule
Altitude & ft \\
Ground speed & kt \\
Track angle & ° \\
5s Ground Speed derivative  & kt/s \\
10s Ground Speed derivative  & kt/s \\
5s track angle derivative  & °/s \\
10s track angle derivative  & °/s \\
\bottomrule
\end{tabular}
\end{table}

%RDC: IMHO this should go to next section
\subsection{Loss and training process}
\label{subsec:loss}

We trained all models on GPUs (RTX A6000 Ada Gen) using Tensorflow with a validation loss checkpoint on the validation set and evaluated the performance of the models on the test set. To ensure reproducibility, we set the same seed in all experiments (ramdom and np.random seed: 42, tf deterministic ops: 1 and python hash seed: 0).

A Binary Cross Entropy (BCE) loss is used to train the model. In addition, we select the best threshold for classification using F1 score on the validation set. Finally, the best model is selected using F1 score, precision and recall on the validation set. The performance of the model is also evaluated on the test set as final performance.

\subsection{Model Architecture}
\label{subsec:model}

\begin{figure*}[ht!]
    \centering
    \includegraphics[width=.99\linewidth]{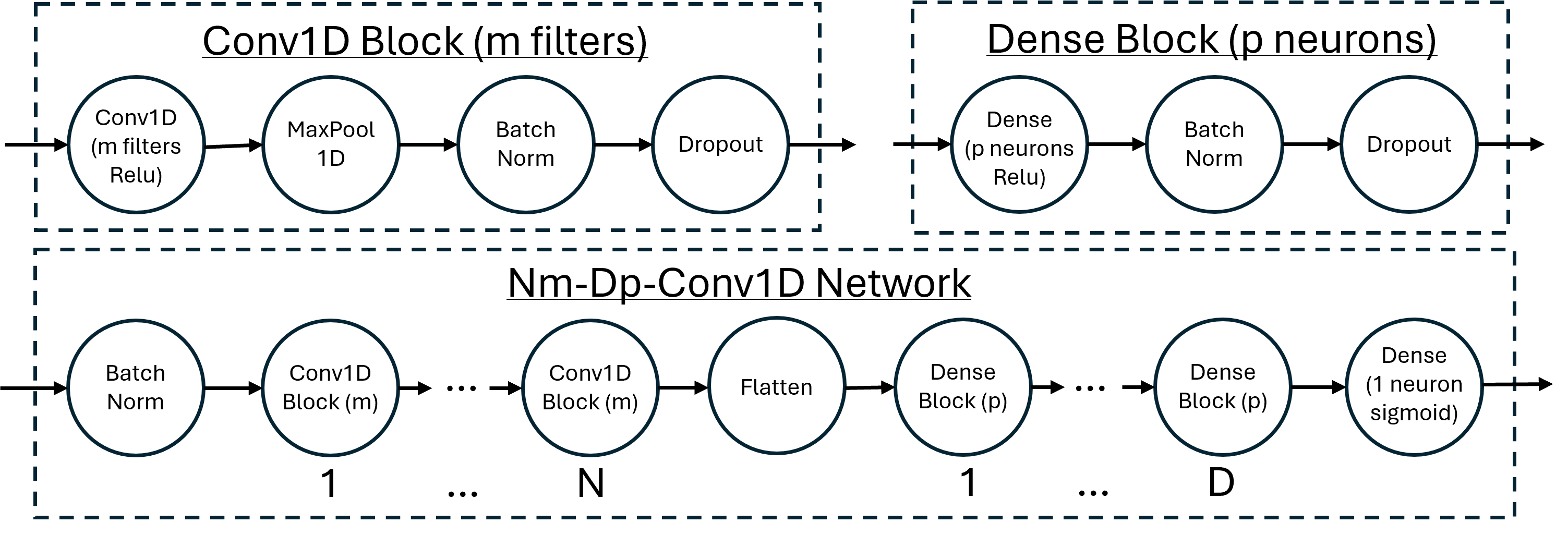}
    \caption{Nm-Dp Conv1D Neural network architecture diagram is displayed at the bottom. On top Conv1D and Dense blocks are displayed.}
    \label{fig:archi}
\end{figure*}

The model we propose for this trajectory classification task is composed of both 1D convolution layer with 1D MaxPooling layers, and Dense blocks. The blocks consist of the respective layer with Relu activation, followed by a batch normalization and a dropout layer.

The proposed architecture (so called Nm-Dp) shown in Figure \ref{fig:archi} starts with a batch normalization, then is composed of N 1D-Convolution blocks with m filters each. Then a flattening layer is used in front of D Dense blocks with p neurons. Finally, the classification layer with one neuron and a sigmoid function gives the output.

\section{Results}
\label{sec:results}

\subsection{Hyper parameter search}

In order to find a good configuration of the model we first launch a fast search to find a starting configuration. Secondly, we continue the analysis by changing one parameter at the time. This different analysis and test are summarized in the following tables of this subsection.

The best learning rate based on validation performance displayed in Table \ref{tab:lr} is 1.0e-04. It achieves the highest F1 score (0.765) and recall (0.812) on the validation set, while maintaining strong precision (0.722). This balance makes it the optimal choice for model performance.

\begin{table}[h!]
\centering
\caption{Performance metrics for different learning rates on validation and test sets.}
\begin{tabular}{ccccccc}
\hline
\multirow{2}{*}{\textbf{Learning Rate}} & \multicolumn{2}{c}{\textbf{F1}} & \multicolumn{2}{c}{\textbf{Precision}} & \multicolumn{2}{c}{\textbf{Recall}} \\
                                     & \textbf{Val}   & \textbf{Test}   & \textbf{Val}   & \textbf{Test}   & \textbf{Val}   & \textbf{Test} \\ \hline

1.0e-03 & 0.717 & 0.67 & 0.733 & 0.689 & 0.702 & 0.653\\
5.0e-04 & 0.709 & 0.588 & 0.678 & 0.611 & 0.744 & 0.567\\
\textbf{1.0e-04} & \textbf{0.765} & \textbf{0.719} & \textbf{0.722} & \textbf{0.667} & \textbf{0.812} & \textbf{0.779}\\
5.0e-05 & 0.738 & 0.623 & 0.689 & 0.533 & 0.795 & 0.75\\
1.0e-05 & 0.507 & 0.407 & 0.6 & 0.544 & 0.439 & 0.325\\
5.0e-06 & 0.293 & 0.278 & 0.367 & 0.344 & 0.244 & 0.233\\

\hline
\end{tabular}

\label{tab:lr}
\end{table}

In this analysis, we fixed the total number of gradient descent runs during training and changed the number of batches per epoch, thus changing the number of epochs. Therefore, here we analyze a combination of epoch number and batch size per epoch. The results summarized in Table \ref{tab:metrics_batch} show that the batch size of 32 offers the best balance. Although larger batch sizes, such as 64 and 128, perform better in precision, their recall values are lower, making batch size 32 the preferred choice due to its superior recall performance, which aligns with our priority.

\begin{table}[h!]
\centering
\caption{Performance metrics for different batch size value on validation and test sets.}
\begin{tabular}{cccccccc}
\hline
\multirow{2}{*}{\textbf{Batch}} & \multirow{2}{*}{\textbf{Epochs}} & \multicolumn{2}{c}{\textbf{F1}} & \multicolumn{2}{c}{\textbf{Precision}} & \multicolumn{2}{c}{\textbf{Recall}} \\
                                    & & \textbf{Val}   & \textbf{Test}   & \textbf{Val}   & \textbf{Test}   & \textbf{Val}   & \textbf{Test} \\ \hline

16 & 50 & 0.687 & 0.582 & 0.633 & 0.511 & 0.75 & 0.676\\
\textbf{32} & \textbf{100} & \textbf{0.765} & \textbf{0.719} & \textbf{0.722} & \textbf{0.667} & \textbf{0.812} & \textbf{0.779}\\
64 & 200 & 0.75 & 0.653 & 0.8 & 0.7 & 0.706 & 0.612\\
128 & 400 & 0.767 & 0.645 & 0.822 & 0.656 & 0.718 & 0.634\\
\hline
\end{tabular}

\label{tab:metrics_batch}
\end{table}

The optimal dropout rate based on validation performance in Table \ref{tab:dropout} is 0.25. It also maintains strong performance on the test set, making it the best choice for regularization while preserving model accuracy and generalization.

\begin{table}[h!]
\centering
\caption{Performance metrics for different dropout rate on validation and test sets.}
\begin{tabular}{ccccccc}
\hline
\multirow{2}{*}{\textbf{Dropout}} & \multicolumn{2}{c}{\textbf{F1}} & \multicolumn{2}{c}{\textbf{Precision}} & \multicolumn{2}{c}{\textbf{Recall}} \\
                                    & \textbf{Val}   & \textbf{Test}   & \textbf{Val}   & \textbf{Test}   & \textbf{Val}   & \textbf{Test} \\ \hline

None & 0.494 & 0.423 & 0.678 & 0.656 & 0.389 & 0.312\\
0.1 & 0.71 & 0.629 & 0.722 & 0.622 & 0.699 & 0.636\\
0.2 & 0.737 & 0.63 & 0.7 & 0.578 & 0.778 & 0.693\\
\textbf{0.25} & \textbf{0.765} & \textbf{0.719} & \textbf{0.722} & \textbf{0.667} & \textbf{0.812} & \textbf{0.779}\\
0.3 & 0.754 & 0.643 & 0.8 & 0.7 & 0.713 & 0.594\\
0.35 & 0.727 & 0.653 & 0.667 & 0.544 & 0.8 & 0.817\\
0.4 & 0.723 & 0.655 & 0.711 & 0.622 & 0.736 & 0.691\\
0.5 & 0.713 & 0.628 & 0.678 & 0.6 & 0.753 & 0.659\\
\hline
\end{tabular}

\label{tab:dropout}
\end{table}

The validation metrics in Table \ref{tab:ND_depth} highlight that a depth of 6 (N and D value of 6) yields the strongest recall and F1 score. While a depth of 4 offers the highest precision at 0.800, its lower recall value of 0.621 makes it less suitable for our focus. Additionally, increasing the depth to 7 results in a slight decline in recall (0.739).

\begin{table}[h!]
\centering
\caption{Performance metrics for different N and D depth value on validation and test sets.}
\begin{tabular}{ccccccc}
\hline
\multirow{2}{*}{\textbf{N \& D value}} & \multicolumn{2}{c}{\textbf{F1}} & \multicolumn{2}{c}{\textbf{Precision}} & \multicolumn{2}{c}{\textbf{Recall}} \\
                                     & \textbf{Val}   & \textbf{Test}   & \textbf{Val}   & \textbf{Test}   & \textbf{Val}   & \textbf{Test} \\ \hline

3 & 0.616 & 0.506 & 0.589 & 0.467 & 0.646 & 0.553\\
4 & 0.699 & 0.577 & 0.8 & 0.711 & 0.621 & 0.485\\
5 & 0.718 & 0.648 & 0.722 & 0.633 & 0.714 & 0.663\\
\textbf{6} & \textbf{0.765} & \textbf{0.719} & \textbf{0.722} & \textbf{0.667} & \textbf{0.812} & \textbf{0.779}\\
7 & 0.747 & 0.694 & 0.756 & 0.667 & 0.739 & 0.723\\

\hline
\end{tabular}

\label{tab:ND_depth}
\end{table}

Result in Table \ref{tab:mp_value} illustrates that using 256 filters and neurons (m and p value of 256) is the best value. While other configurations, such as 128 filters and neurons, offer a decent balance with an F1 score of 0.736 and recall of 0.762, they don't outperform the 256 setting. Increasing the values to 384 or 512 leads to a decline in recall.

\begin{table}[h!]
\centering
\caption{Performance metrics for different m filters and p neurons value on validation and test sets.}
\begin{tabular}{ccccccc}
\hline
\multirow{2}{*}{\textbf{m \& p value}} & \multicolumn{2}{c}{\textbf{F1}} & \multicolumn{2}{c}{\textbf{Precision}} & \multicolumn{2}{c}{\textbf{Recall}} \\
                                     & \textbf{Val}   & \textbf{Test}   & \textbf{Val}   & \textbf{Test}   & \textbf{Val}   & \textbf{Test} \\ \hline

32 & 0.687 & 0.551 & 0.633 & 0.478 & 0.75 & 0.652\\
64 & 0.708 & 0.613 & 0.633 & 0.556 & 0.803 & 0.685\\
128 & 0.736 & 0.682 & 0.711 & 0.667 & 0.762 & 0.698\\
\textbf{256} & \textbf{0.765} & \textbf{0.719} & \textbf{0.722} & \textbf{0.667} & \textbf{0.812} & \textbf{0.779}\\
384 & 0.697 & 0.58 & 0.678 & 0.544 & 0.718 & 0.62\\
512 & 0.749 & 0.652 & 0.744 & 0.667 & 0.753 & 0.638\\

\hline
\end{tabular}

\label{tab:mp_value}
\end{table}

A kernel size of 10 provides the best balance according to the validation results in Table \ref{tab:kernel_size}, which is in line with our goal of optimizing recall. Slightly lower F1 scores (0.762) and recall (0.758) are shown for a kernel size of 5, but the overall performance improves when the kernel size is increased to 10.

\begin{table}[h!]
\centering
\caption{Performance metrics for different kernel size on validation and test sets.}
\begin{tabular}{ccccccc}
\hline
\multirow{2}{*}{\textbf{Kernel Size}} & \multicolumn{2}{c}{\textbf{F1}} & \multicolumn{2}{c}{\textbf{Precision}} & \multicolumn{2}{c}{\textbf{Recall}} \\
                                     & \textbf{Val}   & \textbf{Test}   & \textbf{Val}   & \textbf{Test}   & \textbf{Val}   & \textbf{Test} \\ \hline

5 & 0.762 & 0.671 & 0.767 & 0.622 & 0.758 & 0.727\\
7 & 0.739 & 0.651 & 0.833 & 0.767 & 0.664 & 0.566\\
9 & 0.695 & 0.678 & 0.633 & 0.656 & 0.77 & 0.702\\
\textbf{10} & \textbf{0.765} & \textbf{0.719} & \textbf{0.722} & \textbf{0.667} & \textbf{0.812} & \textbf{0.779}\\
11 & 0.726 & 0.674 & 0.722 & 0.678 & 0.73 & 0.67\\
13 & 0.74 & 0.626 & 0.711 & 0.567 & 0.771 & 0.699\\

\hline
\end{tabular}

\label{tab:kernel_size}
\end{table}

The best performance based on validation metrics is achieved with no class weight adjustment as shown in Table \ref{tab:weight}. It has the highest F1 score (0.765), precision (0.722), and recall (0.812) on the validation set, as well as strong test performance. This suggests that no additional weighting for class 1 provides the most balanced and effective model performance.

\begin{table}[h!]
\centering
\caption{Performance metrics for different class 1 weight values on validation and test sets.}
\begin{tabular}{ccccccc}
\hline
\multirow{2}{*}{\textbf{Class 1 weight}} & \multicolumn{2}{c}{\textbf{F1}} & \multicolumn{2}{c}{\textbf{Precision}} & \multicolumn{2}{c}{\textbf{Recall}} \\
                                     & \textbf{Val}   & \textbf{Test}   & \textbf{Val}   & \textbf{Test}   & \textbf{Val}   & \textbf{Test} \\ \hline
\textbf{No} & \textbf{0.765} & \textbf{0.719} & \textbf{0.722} & \textbf{0.667} & \textbf{0.812} & \textbf{0.779}\\
2.0 & 0.68 & 0.603 & 0.744 & 0.633 & 0.626 & 0.576\\
3.0 & 0.69 & 0.592 & 0.667 & 0.589 & 0.714 & 0.596\\
4.0 & 0.731 & 0.659 & 0.756 & 0.667 & 0.708 & 0.652\\
5.0 & 0.739 & 0.682 & 0.722 & 0.656 & 0.756 & 0.711\\
6.0 & 0.684 & 0.636 & 0.722 & 0.689 & 0.65 & 0.59\\
7.0 & 0.737 & 0.617 & 0.811 & 0.733 & 0.676 & 0.532\\
8.0 & 0.741 & 0.595 & 0.7 & 0.522 & 0.788 & 0.691\\
9.0 & 0.718 & 0.656 & 0.722 & 0.678 & 0.714 & 0.635\\
10.0 & 0.697 & 0.569 & 0.756 & 0.622 & 0.648 & 0.523\\

\hline
\end{tabular}

\label{tab:weight}
\end{table}

\subsection{Best model analysis}

\begin{figure}[ht!]
    \centering
    \includegraphics[width=.99\linewidth]{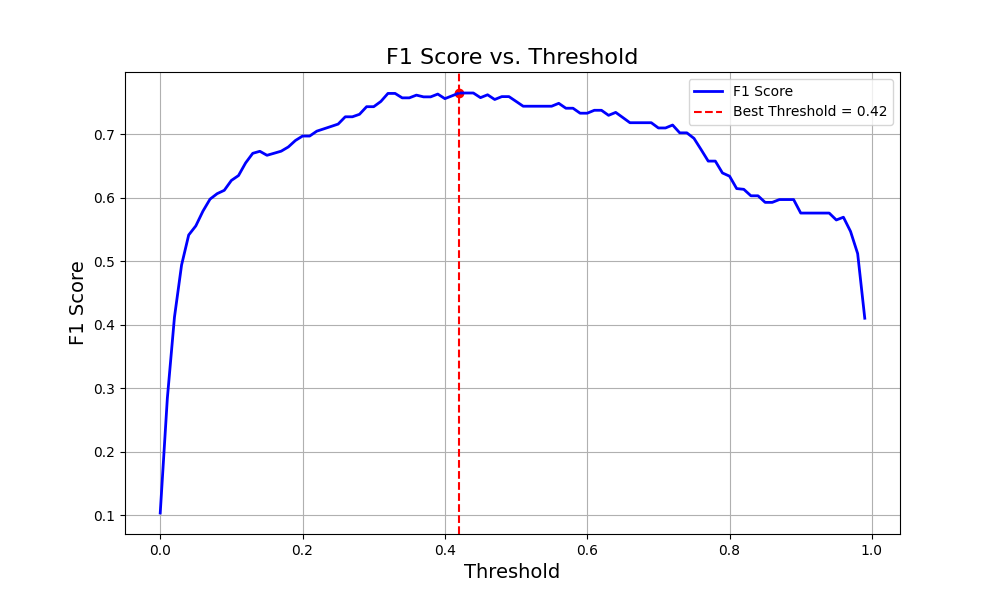}
    \caption{The best classification threshold is computed with F1 score curve on validation set.}
    \label{fig:f1_score}
\end{figure}

In this section, we analyze the best model obtained. First, in Figure \ref{fig:f1_score} we have shown the selection of the classification threshold based on the F1 score. We observe a flat area around 0.3 to 0.5, which corresponds to the optimal threshold to maximize the F1 score (good compromise between precision and recall). The final threshold chosen is 0.42.

\begin{figure}[ht!]
    \centering
    \includegraphics[width=.99\linewidth]{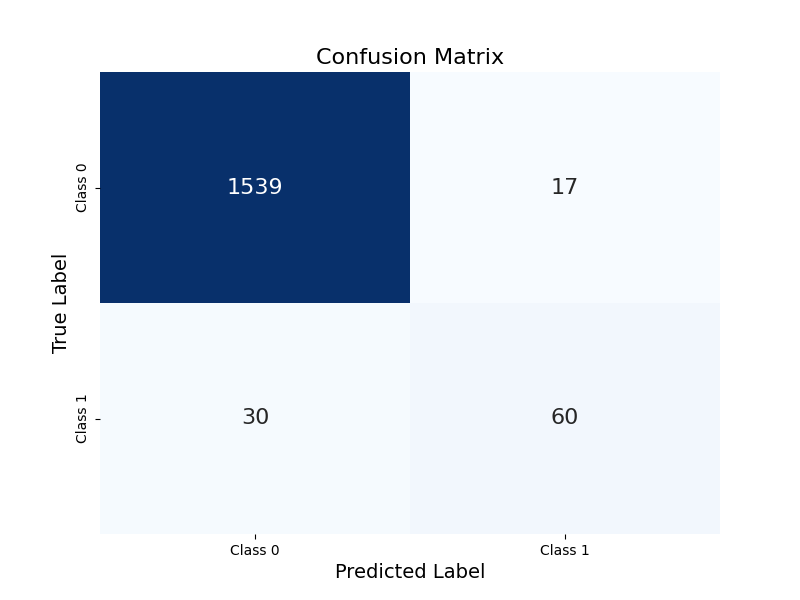}
    \caption{Confusion matrix of the best model on test set.}
    \label{fig:confusion_matrix}
\end{figure}

Additionally, we show the confusion matrix on the test set in Figure \ref{fig:confusion_matrix}. First, we can see that the model predicts very few false positives, which is good behavior. In addition, it manages to predict 60 out of the 90 SETS, therefore, there are 30 false negatives. This is not perfect, but quite good for this difficult task based only on the available ADS-B like parameters. Finally, we expect the model to be conservative, predicting slightly fewer SETS than the ground truth. We will see in the next section how this behaves in terms of environmental impact.

\subsection{Generalization to other aircraft type}

We believe single engine taxiing is less depending on aircraft and engine parameters than fuel flow models in the air and physically link to the value of instaneous thrust addition to accelerate the aircraft. In might not be working for very different aircraft with more engine and with higher mass (ex: A380), however we wanted to validate this hypohesis by applying our best model to another dataset composed of 1910 flight with 42 SETs.

\begin{figure}[ht!]
    \centering
    \includegraphics[width=.99\linewidth]{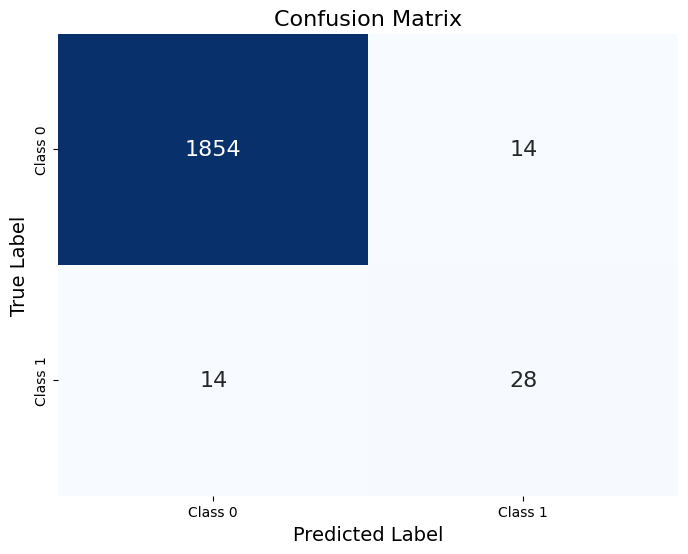}
    \caption{Confusion matrix of the best model on the generalization set composed of 1910 A319 trajectories.}
    \label{fig:confusion_gen}
\end{figure}

The model has an F1 score of 0.676 with a precision of 0.595 and a recall of 0.781. We observe a slight decrease in the performance of the model, as it is still able to categorize SET in a similar proportion as shown in Figure \ref{fig:confusion_gen}, with 2/3 of the SET being correctly labeled. We also observe a very low number of false positives.

\subsection{Localization of the SET start}

In the context of evaluating the fuel flow on ground, a critical part of the SET identification is the proper localization of the engine shutdown. It can be done using predefined cool down period (such as 180sec) after landing, which is supposed to be the nominal operational minimum. We can use a mean value exatracted from train set in our case it would be 255 seconds.

Antoher option consists in building a localization model (regression on the starting index) to properly compute the effective fuel consumption. Such model is tested with the same architecture displayed in Figure \ref{fig:archi}, except for the last layer which is a capped relu activation with maximum of 2047 (size of the padded trajectory minus one since index start at 0). The loss function is a mean square error, and we only build this model using positive samples (real SETs).

A second hyperparameter search is performed, starting with the best features found for the classification task. The parameters changed are the batch size (16), the number of epochs (3000), the dropout (0.6), and the learning rate (5.0e-05).

\begin{figure}[ht!]
    \centering
    \includegraphics[width=.95\linewidth]{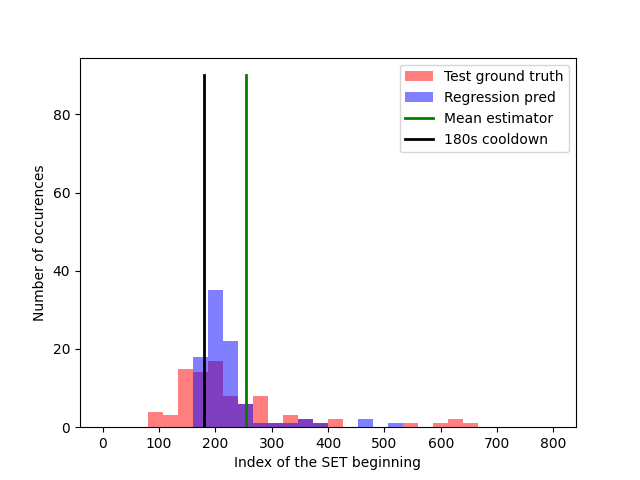}
    \caption{Distribution of SET beginning index and predictions.}
    \label{fig:hist}
\end{figure}

The three methods (180s, mean, and regression model) are compared based on the validation and test set on Mean Absolute Error (MAE) and Mean Error (ME) in Table \ref{tab:reg_perf}. Overall, the task of detecting the start of the set appears to be a very difficult task (best model MAE of 65.65 on test set). We observe that setting a nominal cooldown of 180 sec is the worst case in terms of bias where the regressor is average and the mean estimator the best one (by design). Regarding the MAE, the regressor is the best one on test set with an error of 65.65. Overall when looking at the distribution in Figure \ref{fig:hist}, it confirms the regressor is the best estimator, but that the task is quite challenging.

\begin{table}[h!]
\centering
\caption{Performance of SET starting index estimation for different methods on validation and test sets.}
\begin{tabular}{ccccc}
\hline
\multirow{2}{*}{\textbf{Method}} & \multicolumn{2}{c}{\textbf{MAE Index}} & \multicolumn{2}{c}{\textbf{ME Index}} \\
& \textbf{Val}   & \textbf{Test}   & \textbf{Val}   & \textbf{Test}  \\ \hline

Nominal 180s & 96.3 & 84.9 & 75.05 & 61.74\\
Train mean & 101.74 & 96.88 & 0.31 & -13.31\\
Regression & 100.03 & 65.65 & 32.49 & 19.18\\

\hline
\end{tabular}

\label{tab:reg_perf}
\end{table}

\subsection{Example of fuel consumption calculation}

\begin{figure}[ht!]
    \centering
    \includegraphics[width=.99\linewidth]{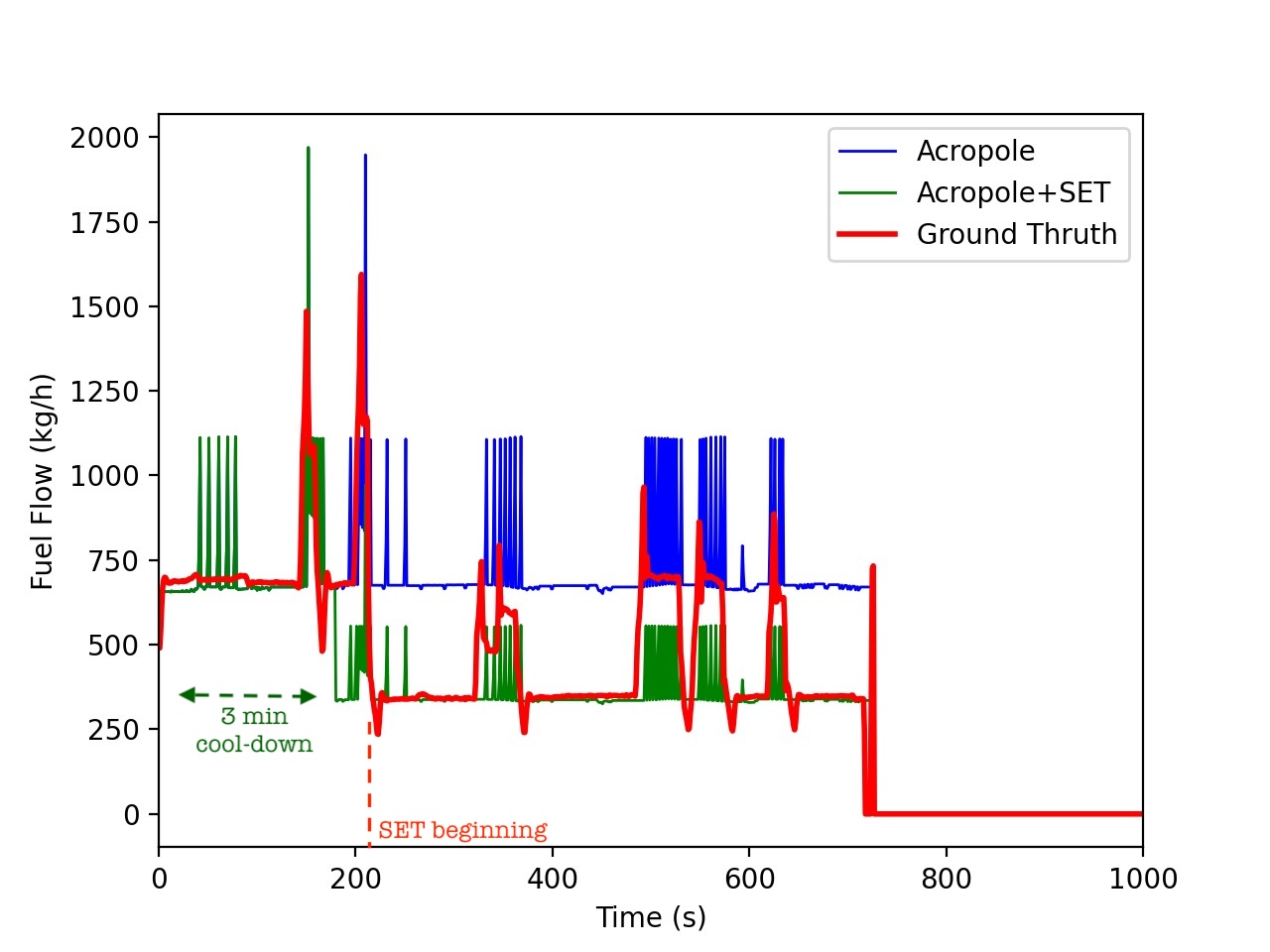}
    \caption{Example of fuel flow estimation and ground truth with and without SET prediction model.}
    \label{fig:SET}
\end{figure}

In this section, we illustrate how this model can be integrated into an environmental impact assessment framework using ground radar data. The process is as follows: first, we apply the Acropole fuel estimation model \cite{jarry2024towards} to the test set cut to the ground taxi-in phase (after aircraft leaves the runway). Second, we apply our single engine taxi detection model. If a trajectory is detected as SET, we consider one of the SET start estimator from previous section and then count only one engine during the remainder of the taxi. In Figure \ref{fig:SET} we show an example result of combining the two models with the nominal 180s cooldown. We see that this particular true positive example shows better estimate consumption by accurately considering a period of SET.

\begin{table}[ht]
\centering
\begin{tabular}{ccccc}
\hline
 \textbf{Model} & \textbf{MAPE} & \textbf{MAE} & \textbf{ME}  \\
 & \textbf{(\%)} & \textbf{(kg)} & \textbf{(kg)}  \\
\hline
No SET & 5.61 (8.26) & 3.89, (9.26) & -0.04 (10.04) \\
SET (Nominal 180s) & 5.00 (6.26) & 3.56 (8.67) & 2.26 (9.1) \\
SET (Train mean) & 4.82 (5.89) & 3.37 (8.02) & 1.93 (8.49) \\
SET (Regression) & 4.85 (5.98) & 3.39 (8.05) & 2.08 (8.48) \\
\hline
\end{tabular}
\caption{Comparison of the estimation of taxiing consumption with or without the set detection for different start estimators.}
\label{tab:perf_set}
\end{table}

As shown in Table \ref{tab:perf_set}, adding the SET detection model to ACROPOLE fuel flow estimation shows superior performance compared to ACROPOLE alone on all key metrics with the three start estimation. With lower MAPE (4.82\% vs. 5.61\%) and MAE (3.37 kg vs. 3.89 kg) for the mean estimator. Indeed introducing SET detection provides more accurate fuel consumption estimates. However, it also increases the Mean Error of around 2kg. This effect is due to false positives. These particular flights induce an underestimation of the fuel consumed since there are badly considered as SET.

To mitigate this problem, the classification threshold was optimized to obtain a good compromise between bias and MAE. A value of 0.9 is obtained (with 2 remaining false positive and 30 SETs detected over 90), which confirms the idea of only predicting SET when the model is sure of it. The results with this threshold are shown in Table \ref{tab:perf_set2}. We observe a reduction of the bias while maintaining a good MAE.

\begin{table}[ht]
\centering
\begin{tabular}{ccccc}
\hline
 \textbf{Model} & \textbf{MAPE} & \textbf{MAE} & \textbf{ME}  \\
 & \textbf{(\%)} & \textbf{(kg)} & \textbf{(kg)}  \\
\hline
No SET & 5.61 (8.26) & 3.89, (9.26) & -0.04 (10.04) \\
SET (0.9, Nominal 180s) & 5.04 (6.59) & 3.43 (7.92) & 1.10 (8.56) \\
SET (0.9, Train mean)  & 5.00 (6.57) & 3.37 (7.70) & 0.96 (8.35) \\
SET (0.9, Regression) & 4.99 (6.55) & 3.37 (7.73) & 1.03 (8.37) \\
\hline
\end{tabular}
\caption{Comparison of the estimation of taxiing consumption with or without the set detection optimised to minimize false positives for different start estimators.}
\label{tab:perf_set2}
\end{table}

\section{Discussion}
\label{subsec:discussion}

The application of the proposed Single Engine Taxi (SET) detection model represents a promising opportunity to improve airport ground operations monitoring by integrating it with real-time data sources. The most reliable data source for implementing this model is ground radar, which provides a high frame rate of one second, ensuring consistent and accurate trajectory tracking. Ground radar's precise coverage allows for real-time monitoring of aircraft taxiing operations, making it an ideal platform for implementing the SET detection model. However, ground radar systems are not universally available at all airports, particularly at smaller or regional facilities, which limits its widespread applicability in such contexts.

A second application of the model could be the use of Automatic Dependent Surveillance-Broadcast (ADS-B) data. While ADS-B is widely used and more readily available at various airports, there are notable limitations in both its coverage and its refresh rate, which typically ranges from 2 to 10 seconds, depending on the aircraft's transponder and the ground station receiving the signals. This lower refresh rate can affect the model's performance in real-time SET detection because the granularity of the data is not as fine as ground radar. In addition, ADS-B coverage may be incomplete in certain areas, particularly at low altitudes or near airport terminals, which could result in gaps in the data needed for accurate model predictions. Further validation using radar or ADS-B data could be performed if QAR data were not anonymized.

The three suggested estimators for detecting the engine shutdown time in the SET model each have distinct strengths and limitations. The fixed 180-second cool-down period is a straightforward approach, but it lacks flexibility, often leading to inaccurate SET starting estimates due to its inability to adjust to real-world conditions. The mean estimator, based on average data from the training set, improves upon this by providing a more balanced approach. However, it still falls short in adapting to specific flight scenarios. The regression model stands out as the most accurate and adaptable method, dynamically adjusting based on real-time data. This results in the lowest mean absolute error (MAE) among the three, but it requires more complex computation and training.

Additionally, the risk of false positives in SET detection can lead to underestimation of fuel consumption, which remains a critical issue, Therefore we recommand to use a high classification threshold. Finally, the regression model’s reliance on high-quality data highlights the need for robust and large data collection systems.

A limitation of the current model is that the data comes from one airline, and to some extent the model may have learned airline specific procedures. An extended dataset with different airlines would be a good way to improve the robustness of the current approach. Finally, We recommend using this SET detection model only for aircraft families with similar engine characteristics, such as the A320 family. The model has been trained on this type, ensuring accuracy in detecting SET. Applying it to aircraft with significantly different configurations, such as wide-body or multi-engine planes, may reduce performance, so its use should be limited to comparable aircraft until further validation.

\section{Conclusions}
\label{sec:conclusion}

In conclusion, this paper presents a novel approach for detecting Single Engine Taxi (SET) operations using deep learning models trained on QAR data. Our approach successfully addresses the challenge of identifying SETs during the taxi-in phase, an operational practice that has the potential to significantly reduce fuel consumption and emissions. Using a convolutional neural network (CNN) architecture, our model achieves an F1 score of 0.765 on the validation set with a recall of 0.812, highlighting its effectiveness in predicting SET with a strong balance between precision and recall. These performance metrics suggest that the model can be reliably used to support sustainability efforts in aviation.

Accurate localization of engine shutdown is critical for SET detection and fuel estimation. Three methods were tested: a fixed 180-second shutdown (MAE: 84.9), a mean estimator from training data (MAE: 96.88), and a regression model (MAE: 65.65). The regression model performed best, although challenges such as false positives remain. To improve accuracy and reduce false positives, a higher classification threshold (0.9) is recommended.

In addition, when such a model is combined with existing fuel consumption estimation tools such as the ACROPOLE model \cite{jarry2024towards}, the overall prediction accuracy is improved. Specifically, our results show that incorporating SET detection reduces the Mean Absolute Percentage Error (MAPE) from 5.61\% to about 5\%. This improvement confirms that SET detection can lead to more accurate taxiing fuel consumption estimates, thereby contributing to greener airport ground operations.

However, there are still opportunities for further research to refine the model. One avenue for future work could be to improve the classification of SET events and the regression of the beginning of the set. Other factors, such as airport or taxiways, could be added as inputs to the model to improve its accuracy. Other models such as Transformer models could be explored. In addition, improving the model's ability to generalize across different aircraft types, particularly those with significantly different configurations or engine counts, such as wide-body aircraft, remains an important goal. This would ensure that the SET detection framework is applicable to a wider range of real-world scenarios. Finally, the models and the implementation for using them are released as an open source library \cite{jarry2024deepenv}.

\section*{Acknowledgement}

The authors would like to thank the Airport Operations \& Innovation team of Orly Airport (ADP) for the discussions on their ongoing work on the identification of single-engine taxiing operations using ground radar data.

\bibliographystyle{ieeetr}

\bibliography{biblio.bib}

\end{document}